\documentclass[sigconf]{acmart}

\copyrightyear{2023}
\acmYear{2023}
\setcopyright{acmlicensed}
\acmConference[CIKM '23]{Proceedings of the 32nd ACM International Conference on Information and Knowledge Management}{October 21--25, 2023}{Birmingham, United Kingdom}
\acmBooktitle{Proceedings of the 32nd ACM International Conference on Information and Knowledge Management (CIKM '23), October 21--25, 2023, Birmingham, United Kingdom}
\acmPrice{15.00}
\acmDOI{10.1145/3583780.3614744}
\acmISBN{979-8-4007-0124-5/23/10}

\usepackage{subcaption}
\usepackage{multirow}

\usepackage{tabularx}
\newcolumntype{C}{>{\centering\arraybackslash}X}
\newcolumntype{Y}[1]{>{\centering\arraybackslash}p{#1}}

\usepackage[inline]{enumitem}

\begin{document}

\title{LARCH: Large Language Model-based Automatic Readme Creation with Heuristics}

\author{Yuta Koreeda}
\email{yuta.koreeda.pb@hitachi.com}
\affiliation{%
    \institution{Hitachi, Ltd., Research and Development Group}
    \streetaddress{1-280 Higashikoigakubo}
    \city{Kokubunji}
    \state{Tokyo}
    \country{Japan}
    \postcode{185-0014}
}

\author{Terufumi Morishita}
\email{terufumi.morishita.wp@hitachi.com}
\affiliation{%
    \institution{Hitachi, Ltd., Research and Development Group}
    \streetaddress{1-280 Higashikoigakubo}
    \city{Kokubunji}
    \state{Tokyo}
    \country{Japan}
    \postcode{185-0014}
}

\author{Osamu Imaichi}
\email{osamu.imaichi.xc@hitachi.com}
\affiliation{%
    \institution{Hitachi, Ltd., Research and Development Group}
    \streetaddress{1-280 Higashikoigakubo}
    \city{Kokubunji}
    \state{Tokyo}
    \country{Japan}
    \postcode{185-0014}
}

\author{Yasuhiro Sogawa}
\email{yasuhiro.sogawa.tp@hitachi.com}
\affiliation{%
    \institution{Hitachi, Ltd., Research and Development Group}
    \streetaddress{1-280 Higashikoigakubo}
    \city{Kokubunji}
    \state{Tokyo}
    \country{Japan}
    \postcode{185-0014}
}


\begin{abstract}
Writing a \emph{readme} is a crucial aspect of software development as it plays a vital role in managing and reusing program code.
Though it is a pain point for many developers, automatically creating one remains a challenge even with the recent advancements in large language models (LLMs), because it requires generating an abstract description from thousands of lines of code.
In this demo paper, we show that LLMs are capable of generating a coherent and factually correct readmes if we can identify a code fragment that is representative of the repository.
Building upon this finding, we developed LARCH (\textbf{L}LM-based \textbf{A}utomatic \textbf{R}eadme \textbf{C}reation with \textbf{H}euristics) which leverages representative code identification with heuristics and weak supervision.
Through human and automated evaluations, we illustrate that LARCH can generate coherent and factually correct readmes in the majority of cases, outperforming a baseline that does not rely on representative code identification.
We have made LARCH open-source and provided a cross-platform Visual Studio Code interface and command-line interface, accessible at \url{https://github.com/hitachi-nlp/larch}.
A demo video showcasing LARCH's capabilities is available at \url{https://youtu.be/ZUKkh5ED-O4}.

\end{abstract}




\maketitle

\begin{figure}[t!]
    \centering
    \includegraphics{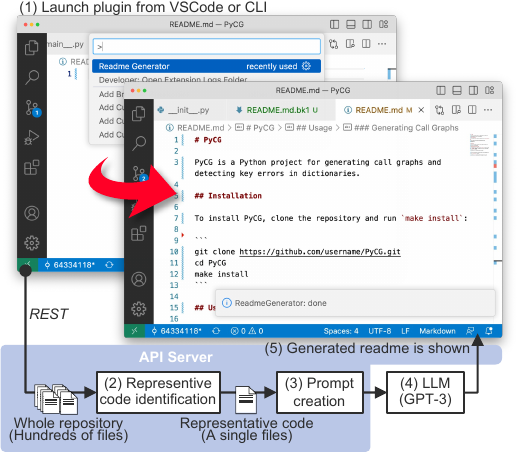}
    \caption{The overview of LARCH (\textbf{L}arge language model-based \textbf{A}utomatic \textbf{R}eadme \textbf{C}reation with \textbf{H}euristics)}\label{fig:system}
    \Description{The overview of LARCH (large language model-based automatic readme creation with heuristics). The figure starts from a screenshot of Visual Studio Code in which the plugin is being brought up. The whole code repository consisting of hundreds of files are being transferred to the representative code identification module. Then, a single file of representative code is extracted from the repository and sent to the prompt creation module. Finally, the prompt is sent to a large language model and the outcome is displayed on Visual Studio Code.}
\end{figure}

\section{Introduction}\label{sec:introduction}

Recent advances in AI, especially large language models (LLMs) \cite{Bommasani2021FoundationModels}, are revolutionalizing software development through code search \cite{husain2019codesearchnet}, program repair \cite{10.5555/3524938.3525939}, code generation \cite{chen2021codex} and many other applications.
However, assisting developers with documentations, which are as important as code itself, is not adequately addressed even though it is a pain point for many developers \cite{10.1145/1085313.1085331}.
In particular, assisting developers write \emph{readme}s is utmost important as it is the most written form of documentation and having no readme essentially makes code unreusable.

The state-of-the-art in assisting developers write a readme is by merely presenting them with a static template or populating the template based on user input, but these solutions do not actually help them write its content.
Previous works have shown that LLMs can generate class-/function-level code comments \cite{wang-etal-2021-codet5,chen2021codex}.
However, comment generation merely involves generating a concrete comment from dozens of lines of code.
Generating a readme remains a challenge as it requires generating an abstract summarization of thousands or even millions of lines of code.

In this demo paper, we show that LLMs are capable of generating a coherent and factually correct readme if we can identify a code fragment that is \emph{representative} (gives overview) of the repository.
Based on this finding, we developed LARCH (\textbf{L}LM-based \textbf{A}utomatic \textbf{R}eadme \textbf{C}reation with \textbf{H}euristics) which is based on representative code identification with heuristics and weak supervision (Figure \ref{fig:system}).
Our contributions are as follows:
\begin{itemize}
    \item We developed LARCH, the first system to generate coherent and factually correct readmes utilizing LLMs.
    \item We show the efficacy of our approach through both human and automated evaluation.
    \item We implemented and open sourced\footnote{\url{https://github.com/hitachi-nlp/larch}} LARCH along with a cross-platform Visual Studio Code (VSCode) interface and command line interface (CLI).
\end{itemize}
In our demo, attendees will have chance to test our system against code of their choice.
You can find the demo video at \url{https://youtu.be/ZUKkh5ED-O4}.

\section{Retrieval-Augmented Language Models}

A language model is a probability distribution over sequences of tokens and it can generate a token sequence by iteratively calculating probabilities of $(i+ 1)$-th token given the \emph{context} of preceding $i$ tokens $\{x_j\}_{j \leq i}$ that were already generated.
In order to generate coherent readme for each repository, we need to carry out generation contextualized by the repository information $\mathcal{R}$.
Following the recent \emph{prompting} paradigm \cite{10.1145/3560815}, we feed the repository information $\mathcal{R}$ as sequences of tokens $\{r_i\}$:
\begin{equation}\label{eq:language_model}
    p(x_0, \dotsc, x_n | \mathcal{R}) = \prod\nolimits_{i=0}^{n-1} p(x_{i + 1} |  r_0, \dotsc, r_{|\mathcal{R}|}, x_0, \dotsc, x_i).
\end{equation}

Recently dominant Transformer-based LMs \cite{NIPS2017_3f5ee243} suffer from a quadric computational complexity against the the sequence lengths.
Since viable context lengths of existing models are much shorter than the average repository size, we need to summarize an input repository to a fixed-size token sequence.
We follow \emph{retrieval-augmented language models} approach \cite{arxiv.2301.12652} and retrieve necessary information from each repository and insert its tokens to $\mathcal{R}$.
Through pilot studies, we found that identifying representative code is the key to readme generation (as demonstrated in Section \ref{sec:experiment}), hence we developed an representative code identification method as described in Section \ref{sec:system-identification}.

\begin{table}[t]
    \centering
    \caption{The labeling functions (LFs) and the features for the entry point identification.}\label{tab:heuristics}
    \fontsize{8pt}{10pt}\selectfont
    \setlength{\tabcolsep}{3pt}
    \renewcommand{\arraystretch}{.7}
    \begin{tabular}{rllcc}\toprule
        \multicolumn{3}{l}{} & \multicolumn{2}{c}{Values for} \\\cmidrule(lr){4-5}
        \multicolumn{3}{c}{Description} & LF\textsuperscript{\textdagger} & Feature \\\midrule
        \multicolumn{5}{l}{\emph{File content}} \\
        & 1. & Contains a string ``main'' in a function name & 1, 0 & 1, 0 \\
        & 2. & Contains an argument parser & 1, 0 & 1, 0 \\
        & 3. & Contains a web framework (such as Flask) & 1, 0 & 1, 0  \\
        & 4a. & Too short ($< 200$ characters) & -1, 0 & --- \\
        & 4b. & Content length (\# characters) & --- & int \\
        \multicolumn{5}{l}{\emph{Directory information}} \rule{0pt}{7pt}\\
        & 5. & Contains substring ``main'' in the file name & 1, 0 &  1, 0\\
        & 6. & Has entry point-ish name (such as ``cli.py'') & 1, 0 & 1, 0 \\
        & 7. & Is ``\_\_init\_\_.py'' & -1, 0 & 1, 0 \\
        & 8. & Has a test-ish name (i.e., starts with ``test\_'') & -1, 0 & 1, 0 \\
        & 9. & Directory depth from the project root & --- & int \\
        \multicolumn{5}{l}{\emph{Static code analysis}} \rule{0pt}{7pt}\\
        & 10a. & Is the top of the import tree & 1, 0 & --- \\
        & 10b. & Distance from the top in the import tree & --- & int \\
        & 11. & Is the bottom of the import tree & -1, 0 & 1, 0 \\
        & 12. & \#  imports (within the repository) & --- & int \\
        & 13. & \#  importers (within the repository) & --- & int \\
        & 14a. & Contains a class inherited $\geq 3$ times & 1, 0 & --- \\
        & 14b. & \#  classes inheriting a class from this file & --- & int \\
        \multicolumn{5}{l}{\emph{Oracle}} \rule{0pt}{7pt}\\
        & 15. & Has the same file name as the repository & 1, 0 & --- \\
        & 16. & Listed as ``entry point'' in ``setup.py'' & 1, -1, 0\textsuperscript{\textdaggerdbl} & --- \\
        & 17. & Imported in the reference readme & 1, -1, 0\textsuperscript{\textdaggerdbl} & --- \\
    \bottomrule
    \addlinespace[1pt]
    \multicolumn{5}{l}{\parbox{236pt}{\fontsize{7pt}{7pt}\selectfont \textdagger 1 if it is likely to be a representative code file, -1 if not, and 0 if it abstains. \\ \textdaggerdbl 0 if it is not 1 and there exists at least one file in the same repository that is 1.}}
    \end{tabular}
\end{table}

\section{LARCH: LLM-based Automatic Readme Creation with Heuristics}

The overview of LARCH is shown in Figure \ref{fig:system}.
Users can launch LARCH from VSCode.
LARCH aggregates code repository from the current workspace, which is sent to the API server.
LARCH identifies the most representative code of the repository using heuristics-based features and gradient boosting trees (Section \ref{sec:system-identification}).
Then, a prompt is constructed from the extracted code (Section \ref{sec:system-prompt}) and it is sent to either an external LLM API or a local LLM (Section \ref{sec:system-llm}).
Finally, the generated readme is sent back to the user and shown on the editor.
The whole process takes about 20 seconds.

The API server comes with CLI which can communicate with API server or generate a readme without launching the server.
They are distributed as a Python pip package and hence cross-platform.
The VSCode interface is distributed as a VSCode plugin (a ``.vsix'' file) and is also cross-platform.

In this paper, we focus on Python projects as it is the most prominent programming language used in the machine learning community.
Nevertheless, our framework can be extended to other languages as well.

\subsection{Weak Supervision for Representative Code Identification}\label{sec:system-identification}

Representative code can take many forms; It can be an entry point of an application, a facade in facade design pattern \cite{gof1994}, or a base class in object-oriented libraries.
Since such complex concept cannot be captured by static program analysis, we propose to employ heuristics-based features that consider diverse properties of a repository, and use machine learning to identify representative code file\footnote{It can be a different granularity such as functions, but we chose files as it is simple and their lengths match LLMs' context lengths.}.

The representative code identification problem has properties that \begin{enumerate*}[label=(\arabic*)]
    \item annotation is quite costly as it requires careful inspection of each repository, and
    \item we can obtain lots of unlabeled public repositories
\end{enumerate*}.
Hence, we decided to take data programming paradigm \cite{10.14778/3157794.3157797}, a weak supervision approach, where we handcraft a set of heuristics to  create silver labels for training a machine learning model.
More specifically, we implemented labeling functions where $j$-th function takes $i$-th file and returns a noisy label ($\in \{-1, 1\}$) or abstains ($= 0$; $\Lambda_{i, j} \in \{-1, 1, 0\}$; $\mathbf{\Lambda} = \{\Lambda_{i, j}\}$).
We can then recover the accuracies of these labeling functions and label posterior for $i$-th file $p(y_i |\mathbf{\Lambda})$ (where $y_i \in \{-1, 1\}$) by solving a matrix completion-style problem\footnote{We used Snorkel (\url{https://www.snorkel.org/}) \cite{10.1609/aaai.v33i01.33014763}.}.

We developed 14 labeling functions (Table \ref{tab:heuristics}).
Notably, we utilize oracle information from reference readmes.
This wouldn't be available in an ordinary weak supervision setting but it should serve as a strong cue for identifying representative code.

\begin{figure}[tb]
    \centering
    \includegraphics{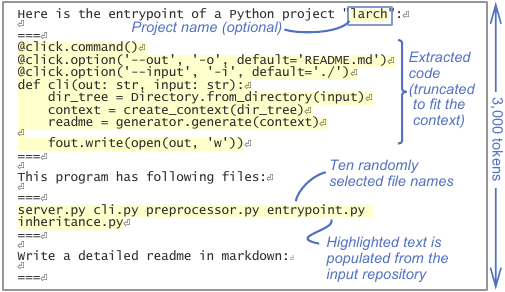}
    \caption{The prompt design}\label{fig:prompt}
    \Description{The figure shows the prompt template. It starts with a phrase ``Here is the entrypoint of a Python project ``larch'':'' which is followed by the extracted representative code. It is followed by text ``This program has following files:'' and ten randomly selected file names. Finally, it contains a phrase ``Write a detailed readme in markdown:'' to mark the start of generation.}
\end{figure}

After obtaining $\{p(y_i = 1 | \mathbf{\Lambda})\}$, we train gradient boosting trees \cite{10.1145/2939672.2939785} to identify representative code files.
We use 14 features that are similar to our labeling functions (Table \ref{tab:heuristics}).
We omit oracle information that wouldn't be available at inference time, and we stop discretizing features to Boolean as we have more flexibility in the values. Instead of formulating the problem as file-wise binary classification, we formulated it as a learning-to-rank problem of files within each repository.
This formulation is more appropriate as our objective is to pick a single file from each repository.
This trick can also improve the overall accuracy by removing repository-level biases of labeling functions (e.g., a repository may contain many files whose names contain ``main'').

\subsection{Prompt Design}\label{sec:system-prompt}

Previous studies have shown that the design of prompt has a significant effect on downstream tasks \cite{10.1145/3411763.3451760,kojima2022large}.
Hence we carefully designed a prompt template that utilizes extracted representative code (Figure \ref{fig:prompt}).
We found that, at least for GPT-3 \cite{ouyang2022training}, imperative sentences are better than declarative sentences (e.g., ``I wrote a readme for this program:'').
Specifying ``markdown'' and ``Python'' did not help much in most cases, but it seems to avoid catastrophic mistakes especially when the input code is short.
If the project name is not given, LLMs generally come up with a name that most matches the code, so we made project name an optional input.
Having file names helps for projects that implement many variations of a single functionality.

\subsection{Large Language Models}\label{sec:system-llm}

We utilized OpenAI API's GPT-3 ``davinci-text-003''\footnote{\url{https://openai.com/api/}} for the LLM.
We utilized prompt length of 3,000 tokens and maximum generation length of 910 tokens.

\begin{table}[t]
    \centering
    \caption{Evaluation of readme generation with representative code identification and the random file baseline}\label{tab:experiment}
        \begin{subtable}[h]{\linewidth}
        \caption{Overall human evaluation}\label{tab:experiment-overall}
        \fontsize{8pt}{10pt}\selectfont
        \renewcommand{\arraystretch}{.6}
        \begin{tabularx}{\linewidth}{lCCC}
            \toprule
            \multicolumn{1}{c}{Context} & Useless & Fair & Useful \\\midrule
            Random file  &  40\% & 20\% & 40\% \\
            Representative code (ours)  & 15\% & 20\% & 65\% \\
            \bottomrule
            \addlinespace[1pt]
            \multicolumn{4}{l}{\parbox{236pt}{\fontsize{7pt}{7pt}\selectfont ``Useful'' when it can be adopted with small fixes, ``Fair'' when it may be useful as a reference, and ``Useless'' otherwise. }}
        \end{tabularx}
    \end{subtable}\\
    \vspace{2ex}
    \begin{subtable}[h]{\linewidth}
        \caption{Fine-grained human evaluation (\% of positive assessments)}\label{tab:experiment-finegrained}
        \fontsize{8pt}{10pt}\selectfont
        \renewcommand{\arraystretch}{.6}
        \begin{tabularx}{\linewidth}{lcc}
            \toprule
            & \multicolumn{2}{c}{Context} \\\cmidrule(lr){2-3}
            \multicolumn{1}{c}{Criteria}  & Random file & Representative code (ours) \\\midrule
            Includes project goal  &  100\% & 100\% \\
            Includes instruction  &  100\% & 100\% \\\midrule
            Grammatical correctness\textsuperscript{\textdagger} &  100\% & 100\% \\
            Markdown correctness\textsuperscript{\textdagger} &  100\% & 100\% \\\midrule
            Factual correctness (text)\textsuperscript{\textdaggerdbl} & 55\% & 75\%  \\
            Factual correctness (code)\textsuperscript{\textdaggerdbl} & 30\% & 65\%  \\
            \bottomrule
            \addlinespace[1pt]
            \multicolumn{3}{l}{\parbox{236pt}{\fontsize{7pt}{7pt}\selectfont \textdagger Percentage of ``Good'' from the choices of ``Bad'', ``Fair'' and ``Good''. \\ \textdaggerdbl ``Good'' when it is mostly correct (e.g., we don't require an example code to run as is, as in a human-written readme), ``Bad'' otherwise.}}
        \end{tabularx}
    \end{subtable}\\
    \vspace{2ex}
    \begin{subtable}[h]{\linewidth}
        \centering
        \caption{Automatic evaluation}\label{tab:experiment-automatic}
        \fontsize{8pt}{10pt}\selectfont
        \renewcommand{\arraystretch}{.6}
        \begin{tabular}{lY{0.8cm}Y{0.8cm}Y{0.8cm}}
            \toprule
            & \multicolumn{3}{c}{ROUGE score $\uparrow$} \\\cmidrule(lr){2-4}
            \multicolumn{1}{c}{Context} & 1 & 2 & L \\\midrule
            Random file  &  20.9 & 4.9 & 10.7 \\
            Representative code (ours)  & 22.0 & 5.9 & 11.4 \\
            \bottomrule
            \addlinespace[1pt]
            \multicolumn{4}{l}{\fontsize{7pt}{7pt}\selectfont Each score considers different $n$-grams. See \cite{lin-2004-rouge} for details. }
        \end{tabular}
    \end{subtable}
\end{table}

\section{Experiments}\label{sec:experiment}

We evaluated LARCH with both human and automatic evaluation.
Since previous studies are limited, we compare LARCH against a baseline that uses a randomly selected Python file instead of representative code, as we believe that representative code identification is the key component of LARCH.

\begin{figure}[tb]
    \centering
    \begin{subfigure}[t]{0.49\linewidth}
        \centering
        \includegraphics{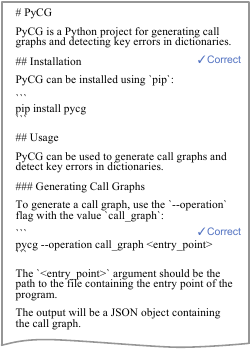}
        \caption{\href{https://github.com/vitsalis/PyCG/tree/64334}{vitsalis/PyCG}}\label{fig:example-pycg}
    \end{subfigure}~
    \begin{subfigure}[t]{0.49\linewidth}
        \centering
        \includegraphics{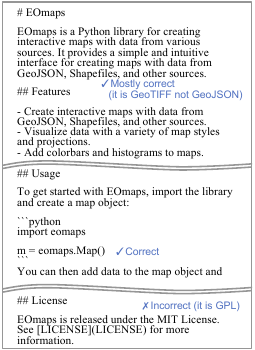}
        \caption{\href{https://github.com/raphaelquast/EOmaps/tree/v4.4.3}{raphaelquast/EOmaps}}\label{fig:example-eomaps}
    \end{subfigure}
    \caption{Excerpts of the generated readmes}\label{fig:example}
    \Description{These set of figures show samples of generated readmes. Both examples show that generated readmes are coherent and mostly factually correct. However, it also points out that LARCH made a minor mistake of writing GeoTIFF as GeoJSON. It also shows that it gets the license term completely incorrect in one example; It says that the repository is MIT but it is actually GPL.}
\end{figure}

We collected public repositories from GitHub\footnote{\url{https://github.com/}}, each of which \begin{enumerate*}[label=(\arabic*)]
    \item has more than 100 stars,
    \item written in Python,
    \item is smaller than 500 MB in size with less than 1,000 files,
    \item contains English readme in markdown format, and
    \item created after the publication of GPT-3 \cite{NEURIPS2020_1457c0d6} (11\textsuperscript{th} June 2020)
\end{enumerate*}.
We randomly sampled 1,500 repositories for automatic evaluation, 20 of which were also used for human evaluations.
We removed a readme and ``setup.py'' from each repository and kept readme as reference data.
Each repository contained, on average, 38.3 files and 103,302 tokens.

For the human evaluation, we compared the two systems in a double-blind setting (pairs of outputs were anonymized and presented in a random order).
The results of the human evaluation are shown in Table \ref{tab:experiment}.
For all the repositories, both systems managed cover both project goals and instructions (``getting started'') and they were coherent both in terms of grammar and markdown formatting.
LARCH, however, performed much better in terms of factual correctness.
This lead to the significant improvement in ``overall usefulness'' ($p=0.009$, Wilcoxon signed-rank test) --- our system performed equally or better than the baseline in 95\% of the repositories.

For the automatic evaluation, we compared ROUGE score \cite{lin-2004-rouge} of generated readmes to the reference readmes, a common metric used in summarization.
As shown in Table \ref{tab:experiment-automatic}, LARCH outperformed the baseline in all metrics.

A couple of example outputs are shown in Figure \ref{fig:example}.
Figure \ref{fig:example-pycg} is a straightforward application and LARCH correctly identified the project goal and its example usage.
LARCH chose \href{https://github.com/vitsalis/PyCG/blob/64334118cf2f2758e1b2e5b972bbce0f46667f44/pycg/__main__.py}{a file with the entry point} where an argument parser and main operations are located.
Figure \ref{fig:example-eomaps} is a class-based library and LARCH chose \href{https://github.com/raphaelquast/EOmaps/blob/v4.4.3/eomaps/_containers.py}{a file with the main class container}.
LARCH got the project features and its usage mostly correct, but got the license completely wrong.
It is natural as we did not incorporate license information to the prompt.
This is an easily amendable problem with some engineerings and we leave it for the future work.

\section{Related Works}

As discussed in Section \ref{sec:introduction}, prevailing approaches to aiding developers in crafting informative readmes predominantly rely on template-based methods.
However, since and around the submission of this paper on June 16, 2023, multiple relevant works have emerged that warrant a discussion.

StarCoder \cite{li2023starcoder} (released on May 4) is the 1.5B parameters state-of-the-art model for code generation.
While readme generation is more of natural language generation than code generation, we believe it is worth running experiments with StarCoder and other code generation models \cite{chen2021codex} in the future work.
There are also multiple new models that support much longer context lengths;
GPT-4-32K \cite{openai2023gpt,openai2023blog} (released on July 6) supports 32K tokens and Claude-2 \cite{anthropic2023model} (released on July 11) supports 100K.
These may lessen the needs for our representative code identification, but we argue that it remains important because \begin{enumerate*}
    \item there exist many repositories that still do not fit onto these larger context lengths,
    \item the performance of LLMs tend to degrade for lengthy inputs even if their positional encodings \emph{support} them \cite{liu2023lost}, and
    \item processing long inputs requires more compute
\end{enumerate*}.
We would nevertheless like to compare and incorporate newer LLMs to our work in the future.

README-AI\footnote{\url{https://github.com/eli64s/readme-ai/tree/v0.0.5}} (version 0.0.1 released on June 28) is a Python library that generates readmes with a LLM but with a slightly different approach.
README-AI summarizes each source file into a short description first, and generates a readme from the concatenated descriptions.
This makes each source file significantly shorter and allows fitting more than one source file to the context.
Nevertheless, we believe that the same arguments as the long context LLMs apply to READM-AI; it does not completely solve the context length limit and it can actually be used along side with our method.
In the future work, we would like to quantitatively compare two approaches and evaluate how they perform if they are put together.

\section{Conclusion}

As presented in Section \ref{sec:experiment}, LARCH can generate coherent and factually correct readme in majority of cases.
We have shown that our representative code identification approach yields much better generation than the baseline.
While there exist risks that LARCH may generate factually incorrect readme, developers can always fix the result.
Since LARCH is straightforward to use and reading readme is much easier than writing one, LARCH can assist developers without having negative effect to the community.

For future work, we will extend our framework to different programming languages.

\section*{Ethical Consideration}

While LARCH significantly improves factual correctness from the baseline, it can still get facts wrong as demonstrated in Section \ref{sec:experiment}.
Misinformation in readmes can have negative effect to the users as it may introduce bugs or causes legal issues with regard to the licencing.
That being said, we intend LARCH to be used by developers of a repository themselves, hence they can always neglect or fix the result.
Since LARCH is straightforward to use and reading readme is much easier than writing one, LARCH can still assist developers without having negative effect to the community.

While we did not run computationally expensive pretraining of LLMs, LARCH still relies on a LLM at the inference which has an unignorable carbon footprint.
Yet, we believe this energy consumption can be justified by resources that we can potentially save by assisting developers.

\begin{acks}
We used computational resource of AI Bridging Cloud Infrastructure (ABCI) provided by the National Institute of Advanced Industrial Science and Technology (AIST) for the experiments.

We would like to thank Dr. Masaaki Shimizu for arranging the computational environment and Takuo Shigetani for helping us with the UI implementation.
\end{acks}

\bibliographystyle{ACM-Reference-Format}
\balance
\bibliography{paper-2023a}


\end{document}